*CORRESPONDENCE
Muhammad Burhan Hafez
✉ burhan.hafez@uni-hamburg.de

†These authors share first authorship

# Map-based experience replay: a memory-efficient solution to catastrophic forgetting in reinforcement learning

Muhammad Burhan Hafez*†, Tilman Immisch†, Tom Weber and Stefan Wermter

Knowledge Technology Research Group, Department of Informatics, University of Hamburg, Hamburg, Germany

Deep reinforcement learning (RL) agents often suffer from catastrophic forgetting, forgetting previously found solutions in parts of the input space when training new data. Replay memories are a common solution to the problem by decorrelating and shuffling old and new training samples. They naively store state transitions as they arrive, without regard for redundancy. We introduce a novel cognitive-inspired replay memory approach based on the Grow-When-Required (GWR) self-organizing network, which resembles a map-based mental model of the world. Our approach organizes stored transitions into a concise environment-model-like network of state nodes and transition edges, merging similar samples to reduce the memory size and increase pair-wise distance among samples, which increases the relevancy of each sample. Overall, our study shows that map-based experience replay allows for significant memory reduction with only small decreases in performance.

## 1. Introduction

Reinforcement learning (RL) has seen a lot of major advances over the last years. This includes solving the game of Go (Silver et al., 2016) and other challenging games (Silver et al., 2018; Schrittwieser et al., 2020), as well as achieving breakthroughs in robotic manipulation (Levine et al., 2016), locomotion (Haarnoja et al., 2019), and navigation (Zhu et al., 2017). Newer algorithms, such as *AplhaZero* (Silver et al., 2018) and *MuZero* (Schrittwieser et al., 2020), have become more flexible and less dependent on expert human knowledge, which can introduce biases into the learning process and be come expensive to acquire. Furthermore, it has been shown that superhuman artificial intelligence programs built using these newer algorithms contribute to improvements in human decision-making by encouraging novelty in human decisions (Shin et al., 2023). A major part of these advances has been the introduction of deep function approximators based on neural networks (NNs) (Mnih et al., 2015), well suited to learning complex functions over massive continuous input spaces, into RL. They are, however, prone to catastrophic forgetting (McCloskey and Cohen, 1989), a phenomenon where the network encounters a sudden and drastic drop in performance in known scenarios while learning novel information. This is an especially challenging problem in RL, because catastrophic forgetting is exacerbated by long sequences of highly correlated data, which is the default for any RL task. If all incoming batches of data are always highly correlated, changing only by slight variations, then an agent could achieve a high prediction accuracy by continuously adjusting to its immediate environment while forgetting older experiences, because the latter probably is insignificant in decision making in the current surroundings. If such an agent was now exposed to some part of the state space different from the one it was most recently trained on, it would not know how to act.

Replay buffers are a popular method to mitigate catastrophic forgetting (Riemer et al., 2019; Rolnick et al., 2019; Daniels et al., 2022), as they can store and periodically retrain on inputs that are very rare or lie far in the past, but that may become relevant again when the input distribution changes or when generalizing to new situations by learning from past experiences. This is because over time the input space might change in some way that makes previously rare inputs more common or relevant. For instance, the NN might have been trained on data collected under certain conditions, but later needs to work in a different context where previously rare inputs are now more frequent. If the training data for the NN is diverse, comprising both old and new samples from all over the input space, the model has no choice but to adapt to all of them to reduce the overall error. Traditionally, these replay memories merely store something like (state, action, next-state, reward)-tuples at every training step to construct a dataset so that the agent can then batch-learn from it. Storing every one of these tuples can be storage intensive and inefficient, not taking into account the similarity of tuples, thus possibly storing some that are virtually identical or only different to irrelevant degrees. This will also increase the chances of training batches being more correlated, that is, containing a high proportion of such samples that are irrelevantly dissimilar.

Our approach, the Grow-When-Required-Replay (GWR-R), uses a growing self-organizing network based on the Gamma-Grow-When-Required (GGWR) algorithm (Parisi et al., 2017) that maps an input space to a compact network. We merge similar states into nodes and associate actions and rewards to directed network edges. While this causes long computation times as additional calculations have to be performed that are not necessary for more straightforward storage methods, this approach can reduce memory sizes significantly with disproportionally small decreases in performance. As similar inputs are merged, we suspect that diversity, meaning overall pairwise distances, of the memory samples is increased. This theoretically leads to more diverse training batches. Our approach is similar to a model-based RL (Schrittwieser et al., 2020; Hansen et al., 2022), where the agent learns a transition model of the environment and uses it to sample synthetic transitions for training the policy and the value function with higher sample efficiency.

We conduct a series of experiments regarding the influence of two hyperparameters of the GWR-R on performance, namely, the activation threshold $a_T$ and the habituation threshold $h_T$, that both influence the compression of samples into fewer nodes. Experiments are carried out by running the DDPG algorithm (Lillicrap et al., 2016) on the continuous Multi-Joint-Controller (MuJoCo) environments Inverted-Pendulum, Reacher, HalfCheetah, and Walker2D (Todorov et al., 2012) and comparing the results to the standard DDPG replay buffer. We find that, by lowering the activation threshold, considerable memory compression is possible, with only small losses in performance. An implementation of our GWR-R approach can be found at https://github.com/TilmanImmisch/GWRR.

The primary contributions of our work are summarized as follows:

- We develop a cognitive architecture that resembles a mental model of the world for simulating experiences in RL.
- We present a novel experience replay approach to address catastrophic forgetting in RL that is more memory-efficient than standard experience replay, supports state abstraction, and simulates state transitions.
- We show that our approach significantly reduces the amount of distinctly stored samples and achieves memory reduction of 40−80% compared to an established RL baseline with standard experience replay while retaining comparable performances across four simulated robot control tasks (Inverted-Pendulum, Reacher, HalfCheetah, and Walker2D).

## 2. Related work

The search for better replay memories and sampling methods to train Deep RL agents is a popular topic as experience replay (ER) presents an efficient and simple solution to catastrophic forgetting. Improved methods promise potentially faster and more stable convergence with smaller amounts of samples needed to be stored.

### 2.1. Selective sampling and training

We give a quick overview of methods of selective sampling where the developed methods store all transitions encountered, but deviate from standard uniform sampling methods. Daley et al. (2021) proposed a *Stratified Experience Replay* (SER) to counteract remaining correlation when sampling uniformly that uses a stratified sampling scheme to efficiently implement a non-uniform sampling distribution to correct for multiplicity bias, which leads to much faster learning in small Markov Decision Processes (MDPs).

The *Remember and Forget Experience Replay (ReF-ER)* (Novati and Koumoutsakos, 2019) addresses false policy gradients caused by changes in sample distributions as the policy changes. It calculates the significant weights of all experiences by computing the ratio of probability to choose a past action with the new policy vs. the old (off-policy) behavior, clipping gradients to 0 when the ratio is too large or too small. It achieves strong results in many continuous environments by penalizing policy changes to stay more in line with replay behavior.

By averaging overall stored rewards for a given state-action pair and attaching all possible next states, von Pilchau et al. (2020) generated synthetic samples for stabilizing learning in non-deterministic environments. This way synthetic experiences present a more accurate estimate of the expected long-term return of a state-action pair, boosting performance in comparison to a standard ER.

Using the *temporal difference error* (TD-error) as a substitute measure for learning potential, Schaul et al. (2015) prioritized sampling high TD transitions. The last computed TD-error is stored along with each transition, and mini-batches of samples are then selected stochastically, increasing the probabilities of high TD and just encountered samples and leading to more efficient learning.

## 2.2. Selective storage

In selective storage methods, the way the samples are stored is different from standard approaches, that is, not storing full transition information or skipping out on some transitions entirely. Isele and Cosgun (2018) explored four measures for evaluating which samples to store, namely, (1) surprise, favoring high TD samples; (2) reward, favoring high expected return samples; (3) distribution matching, which matches the global stored state distribution; and (4) coverage maximization, which attempts to cover the state space as fully and evenly as possible. In all (multi-task) experiments, except in cases where tasks which received limited training were more important, distribution matching was the top performer, the exception was coverage maximization.

Using NNs, Li et al. (2021) learned a model of the environment to reduce memory needs in a dual replay buffer architecture. They trained Deep Q-Network (DQN) on both real data from a short-term replay buffer and generated data from the *Self-generated Long-Term Experience Replay*, in which they store just episode start states and all following actions and rewards, letting intermediate states be inferred by their environment model. This way they achieve comparable performance on multi-task continual environments, such as on Star Craft II (Vinyals et al., 2019), alleviating catastrophic forgetting while saving on memory.

Another double replay approach (Zhang et al., 2019) employs an exploration and an exploitation buffer, modulating the ratios of samples between the two over the course of training. While the former retains samples by reservoir sampling (Vitter, 1985), storing and deleting samples by probability based on buffer size, the latter uses a standard first-in-first-out (FIFO) scheme. The selection ratio of exploration samples is then modulated based on the maximum of either the exploration parameter $\epsilon$ or another measure $\tau$ that increases as the policy target and in-use network stabilize. This approach results in improved training and generalization performance.

In GWR-Replay (GWR-R) approach, state abstraction is supported by the growing self-organizing network and allows for simulating transitions without the need for replay buffers. None of the discussed selective sampling and selective storage approaches are based on state abstraction or they avoid using replay buffers.

## 2.3. Developmental robotics

Growing self-organizing networks have recently been used to support the research on developmental robotics as they enable the continual learning of both representations and skills. The instantaneous topological map (Jockusch and Ritter, 1999), a type of growing self-organizing network, was used to self-organize the sensory space into local regions. A local world model is assigned to each region, and the model's learning progress derives an intrinsic reward to encourage directed exploration for vision-based grasp learning on a developmental humanoid robot (Hafez et al., 2019a,b). The self-organizing network is not used to simulate experiences but a replay buffer with a large and fixed capacity is used instead. In a different study, GWR was used to learn separate unimodal mappings for sensory and motor information in gaze and arm control and to learn multimodal associations between the mappings (Rahrakhshan et al., 2022). While this was shown to enable learning robotic eye-hand coordination skills, it required collecting a large dataset of sensory-motor pairs for learning and did not consider using GWR for memory-efficient RL. In addition to learning single tasks, GWR was also used for learning a behavior embedding space from unlabeled task demonstrations for continual multi-task robot learning (Hafez and Wermter, 2021), but disregarding memory efficiency. These works were evaluated on our Neuro-Inspired COmpanion (NICO) robot (Kerzel et al., 2017), a child-sized humanoid for research on embodied neurocognitive models based on human-like sensory and motor capabilities.

Our map-based experience replay approach presented in this study strongly assists the above-mentioned research on developmental robotics by providing a simple, cognitive-inspired, and memory-efficient solution to catastrophic forgetting. We suggest that our approach can be particularly useful for the developmental and simultaneous learning of high-level world models and skills as it supports state abstraction and simulates state transitions inside a growing self-organizing network. Similarly, our approach has a high potential for equipping our NICO robot with continual learning abilities necessary for acquiring a diverse set of skills.

# 3. Approach

We propose a self-organizing network experience replay based on G-GWR (Parisi et al., 2017), called GWR-R, which is similar to model-based RL approaches. The GWR-R represents the movement of the agent through the input space as a directed graph connecting states with directed edges of actions (and rewards), the direction being forward in time. It merges seen states that are very similar to each other into network nodes, reducing the size of the sampling set in such a way that all sample experience tuples are different from each other by a degree set by tuning the GWR-R.

The GWR-R approach abstracts and generalizes over the state space of an RL agent by mapping input states to the closest node in the network. Since the distance between two temporally adjacent nodes in the resulting map, which is the graphical representation of the data, is larger than the distance between any two states that belong to either node, we theorize that this approach creates a more diverse set of samples. Thus each batch the RL agent samples and trains on should on average be more diverse than a batch from a classical replay buffer, making the agent relearn more diverse parts of the state space each iteration, which should reduce catastrophic forgetting overall and lead to more stable performance. However, the methods by which nodes and edges are constructed can make them deviate from reality, perhaps by creating tuples that are impossible or improbable in the real environment, which would impact learning negatively. We retain the structure of inputs and outputs the same as in the DDPG's replay buffer, making it easy to attach to many state-of-the-art RL algorithms that use a replay buffer.

## 4. GWR-replay

The GWR-R is map-based experience replay, because it replays experiences in the map space rather than the state space during training. It functions very similarly to the GGWR (Parisi et al., 2017). The GWR-R is initialized with two nodes in random positions in the map space. The map space is set to have the same dimensions as the state space of the environment the agent lives in. Like most standard replay memories, the GWR-R receives an input of tuples of ($s_t, s'_t, a_t, r_t$, and $d_t$) at time step $t$, $s_t$ being the previous state the agent was in, $s'_t$ being the state the agent is in now, after having performed an action $a_t$, receiving a reward $r_t$, and binary information $d_t$ regarding whether the episode is now completed or not. Each state is input into the GWR-R as a sample x(t), after which it selects the respective best-matching unit (BMU), that is, the closest existing node to x(t) in the map space, and either updates the BMU or creates a new node, depending on the condition $a_b < a_T \wedge h_b < h_T$, where $a_T$ and $h_T$ are the activation and habituation thresholds, $a_b$ $h_b$ are the activation and habituation of the BMU. The activation $a_b$ is a measure of the distance *dist* from BMU to x(t), computed as $a_b = e^{(-dist)}$, and the habituation $h_b$ is a counter storing the number of times a node has been selected as BMU, but in a descending logarithmic means from 1 to 0, thus it can also be used to decrease the learning rate with time. Differing from the original implementation, when a new node $n$ is to be inserted, it is not inserted halfway between the BMU and $x(t)$, but exactly at the position of $x(t)$. That way, the nodes in the GWR-R represent the states the agent sequentially visits.

We consider the state $s_t$ in the first experience tuple of the current episode and the next state $s'_t$ in the last tuple of the previous episode as two different (temporally disconnected) samples in two sequential training iterations, while in the middle of an episode, as $s'_{t-1} \equiv s_t$, usually, per tuple we just input $s'_t$ into the GWR-R to be learned. To account for the rest of the tuple, the done flag $d_T$ is associated with either the BMU $b_{s'_t}$, closest to $s'_t$, or when a new node is created, to that new node. It is integrated with an existing done flag through an exponentially weighted average

$$d_{b_{s'_t}}(n) \leftarrow \phi \cdot d_t + (1 - \phi) \cdot d_{b_{s'_t}}(n-1) \tag{1}$$

and for a new node simply set to $d_t$

$$d_{b_{s'_t}}(1) = d_t, \tag{2}$$

where $n$ is the number of times a node has been activated and trained and $\phi$ is a weight coefficient controlling the influence of the current sample on the existing average.

For processing the action $a_t$ and the reward $r_t$, we use directed temporal edges, which extend on the temporal connections introduced in Parisi et al. (2016). Similar to the GGWR, we create a temporal count matrix $T^C$ with dimensions (network size x network size) that simply counts how often two network nodes have been activated sequentially as BMUs. It does so by setting

$$T^C_{\{b_{s_t}, b_{s'_t}\}} + = 1, \tag{3}$$

each time $b_{s'_t}$ comes after $b_{s_t}$, with all entries being initialized as 0. In this case, two consecutively activated BMUs represent two states the agent experiences one after the other, $T^C$ thus counting the number of times a particular state was followed by another.

We additionally introduce two new temporal matrices for action and reward, $T^A$ and $T^R$, which store the action $a_t$ and reward $r_t$ as directed edges between $b_{s_t}$ and $b_{s'_t}$. $a_t$ and $r_t$ are also averaged into existing values by an exponentially weighted average as follows:

$$T^A_{\{b_{s_t}, b_{s'_t}\}}(j) \leftarrow \phi \cdot a_t + (1 - \phi) \cdot T^A_{\{b_{s_t}, b_{s'_t}\}}(j-1), \tag{4}$$

$$T^R_{\{b_{s_t}, b_{s'_t}\}}(j) \leftarrow \phi \cdot r_t + (1 - \phi) \cdot T^R_{\{b_{s_t}, b_{s'_t}\}}(j-1), \tag{5}$$

setting $T^A_{\{b_{s_t}, b_{s'_t}\}}(1) \leftarrow a_t$ and $T^R_{\{b_{s_t}, b_{s'_t}\}}(1) \leftarrow r_t$, where $j$ is the number of times an edge has been trained and $\phi$ is the same weight as in Equation 1. Overall, we define a temporal edge (directed) to consist of the three values of the count, average action, and average reward.

Each time an episode in the RL environment starts, the first state is learned without a temporal connection to the preceding BMU, in this case, the final state of the previous episode. Processing tuples like this results in long temporal trajectories of BMU activations in the GWR-R representing episodes. We now have two different kinds of network edges, the undirected *neighborhood edges* (Parisi et al., 2017) that connect nodes that were activated at the same time as BMU and second-BMU and the newly introduced *temporal edges* in $T$ that connect nodes that were activated consecutively as BMUs. An illustration of the GWR-R is shown in Figure 1.

After training the GWR-R, to sample a tuple of shape $(s, s', a, r, d)$ for training an RL agent, a random node in the network is selected to represent $s$. Then, a follow-up candidate temporally connected to $s$ is selected from all non-zero entries $I$ in $T^C_{\{s,*\}}$, where the probability for choosing $s'$ is $Pr[s'] = \frac{T^C_{\{s,s'\}}}{\sum_{i \in I} T^C_{\{s,i\}}}$, retrieving the done flag of $s'$, $d \leftarrow d'_s$, and the action and reward stored along the edge as $a \leftarrow T^A_{\{s,s'\}}$ and $r \leftarrow T^R_{\{s,s'\}}$.

When setting $a_T$ and $h_T$ as very high, the GWR-R inserts a node for every sample, exactly mimicking the original replay memory. Lowering either threshold increases the number of samples that are merged into existing nodes, thereby increasing abstraction. The complete algorithm for training an RL agent with the GWR-R is given in the Algorithm 1.

## 5. Experiments

We perform several experiments with our approach to explore how far abstraction can go while maintaining reasonable performance on different tasks and to test our hypothesis that the proposed method may also stabilize learning. For all experiments, every parameter configuration is run multiple times and averaged, as the training of RL agents is stochastic and results can differ

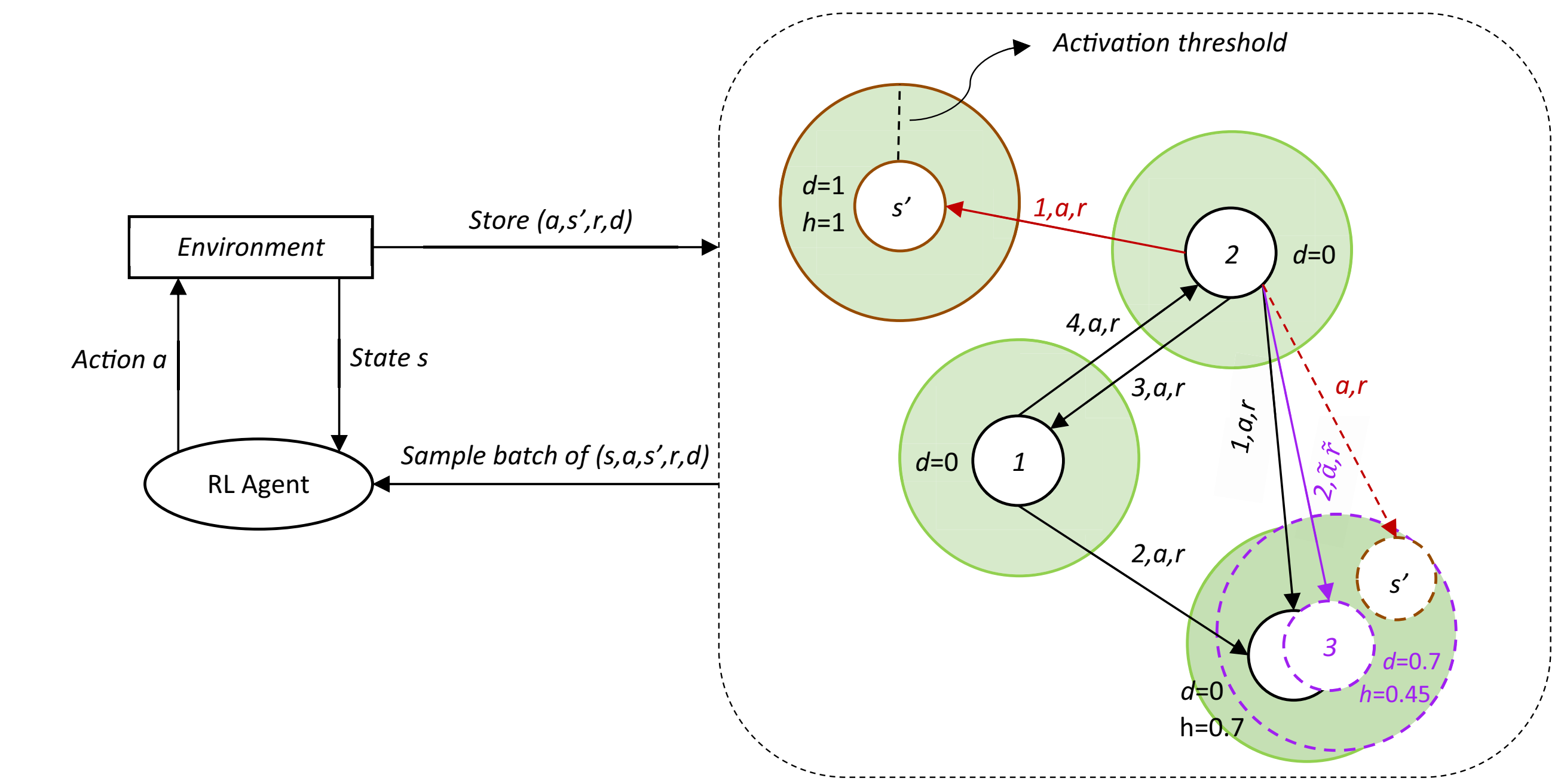

FIGURE 1
A simplified representation of the GWR-R. The RL agent uses the DDPG algorithm and is placed into some environment, which is initialized at some start state $s$. The DDPG responds to that state with an action $a$ it thinks is most appropriate, to which in turn the environment responds and returns a new state $s'$, along with the reward $r$ for that state and whether or not it is a final state as binary flag $d$. $s'$ is again put into the DDPG for it to react, along with the prior action $a$, $r$, and $d$ are stored in the GWR-R. The GWR-R stores the states as nodes (white circles), each with an associated probability of being final $d$ and a habituation counter $h$. Action $a$ and reward $r$ are stored as temporal edges with an associated count of how often the edge has been updated. Two scenarios are shown, one where the input $s'$ (top-center) is outside of the BMU's activation threshold (BMU in this scenario is the white circle with number 2 inside) and is inserted as a new node (circle with a solid brown line) with habituation $h = 1$, and one where the input $s'$ (bottom-right) is inside and the BMU (bottom-right circle with a solid black line) learns toward the input, moving to the new position (circle with a dotted purple line), decreasing its habituation and averaging the action/reward vectors (purple $\tilde{a}$ and $\tilde{r}$) and the done flags. In this second scenario, $s'$ is discarded after the BMU has adjusted position. Note that through this procedure, the resulting average action vectors are not guaranteed to directly point at the position of the associated target node in the environment.

```
s = env.reset()
for t=0, t < max_timesteps, t+1 do
  if t < start_timesteps then
    a = env.action_space.sample()
  else
    a = policy.select_action(s)
  end if
  (s',r,d) = env.step(a)
  gwrr.add(s',a,r,d) (see Algorithm 2)
  if t ≥ start_timesteps then
    samples = gwrr.sample(batch_size)
    policy.train(samples)
  end if
  s = s'
  if d==1 then
    s = env.reset()
    gwrr.add(s) (see Algorithm 2)
  end if
end for
```

**Algorithm 1.** Training of an RL-agent with the GWR-R.

substantially from run to run. We modulate the thresholds $a_T$ and $h_T$ that we deem, especially, relevant to our approach. Hyperparameter values that are not explicitly stated are set to the values in Table 1. We compare our approach to a baseline, focusing on the performance and memory size.

## 5.1. Setup

Our RL algorithm of choice is the open source implementation of DDPG included in the repository of Fujimoto et al. (2018) with the configuration of the original DDPG paper (Lillicrap et al., 2016). The hyperparameters are listed in Table 1. Experiments were conducted using the MuJoCo Library v. 2.0.0 (Todorov et al., 2012) from DeepMind, because it supports continuous control with accurate and efficient physics simulation. We selected the environments InvertedPendulum, Reacher, and HalfCheetah, which have 4-, 11- and 17-dimensional state spaces and 1-, 2- and 6-dimensional action spaces respectively, to see possible changes in performance with increasing problem complexities. More experiments are detailed in Appendix A. To get indicative

```
1: Start with two random neurons
2: Initialize set of neighborhood edges and global
   context
3: [GWR-REPLAY] Initialize three empty sets of
   temporal connections T^C = Ø, T^A = Ø and T^R = Ø
   for storing Count, Action and Reward.
4: [GWR-REPLAY] Receive an input sample (s'_t, a_t, r_t, d_t)
5: Select BMU (b_t) and second-BMU (sb_t)
6: if a_{b_t} < a_T and h_{b_t} ≤ h_T then
7:   a: Add a new neuron k at position s'_t with h_k = 1
8:   b: Create neighborhood edges to BMUs
9:   c: [GWR-REPLAY] Create a temporal connection by
     T = T ∪ {b_{t-1}, b_t} for T^A, T^R, T^C and set T^A_{b_{t-1},b_t} = a_t,
     T^R_{b_{t-1},b_t} = r_t and T^C_{b_{t-1},b_t} = 1 and set d_{b_t} = d_t
10: else
11:   a: Create or reset neighborhood edge between
      BMUs
12:   b: Update winning neuron and its neighbors
13:   c: [GWR-REPLAY] Increment T^C and update T^A, T^R
      and d_{b_t} by Equations 4, 5 and Equation 1
14:   Increment age of neighborhood edges, remove old
      edges and neurons with no edges left
15:   Reduce h_{b_t} and h of neighbors
16: end if
17: Update global context
18: If the stop criterion is not met, repeat from
    step 4.
```

**Algorithm 2.** The GWR-R algorithm in contrast to the standard GWR training.

results, all selected parameter configurations were run a total of 10 times, for 100,000 time steps, using the ray-tune library (Liaw et al., 2018) (It is chosen because it supports distributed hyperparameter tuning.), and then averaged.

We tracked the performance by running evaluation episodes at fixed intervals to evade the Gaussian noise added to each action (the policy output) during normal training for some random exploration of the environment. At every evaluation step, six evaluation episodes were completed and their scores were averaged. We compared our approach to the DDPG with its standard replay buffer as the baseline, which simply stores a tuple at each time step in a table and returns a batch of randomly sampled table rows for training. The standard replay buffer stores a maximum of 1,000,000 samples and starts replacing the oldest samples on reaching that threshold by a FIFO scheme. We also tracked memory size, which is the amount of stored non-zero rows in the baseline and the number of nodes in the GWR-R. This might seem unfair at first, as nodes in the GWR-R correspond to a single state and samples in the baseline to two states $(s, s')$, but for each input tuple action, reward and done are processed in the same proportion as with the original, so we chose to keep these definitions of memory size as they provide for very intuitive comparison.

InvertedPendulum experiments were run on a server with two GeForce RTX 2080TI GPUs and an Intel(R) Xeon(R) CPU E5-2620 v4 @ 2.10GHz. HalfCheetah and Reacher were run on different servers with the same GPU setup, but equipped with an AMD EPYC 7551P 32-Core Processor CPU. We always ran two trials in parallel. Each experiment took approximately 2-6 days to complete, as shown in Tables 2–4 for runtimes.

**TABLE 1** Default hyperparameters for experiments.

| | Hyperparameters | Values |
|---|---|---|
| GWR-R | Activation threshold | $a_T = 0.95$ |
| | Habituation threshold | $h_T = 0.8$ |
| | Habituation count | $\tau_B = 0.3, \tau_N = 0.1, \kappa = 1.05$ |
| | Edge averaging weight | $\phi = 0.7$ |
| | No. of context descriptors | $K = 4$ |
| | Influence of context descriptor | $\beta = 0.7$ |
| | Maximum age (Edge) | $age_{max} = 10$ |
| | Learning rates | $\epsilon_B = 0.1, \epsilon_N = .001$ |
| DDPG | Exploration noise | 0.1 (Std of Gaussian) |
| | Batch size | 64 |
| | Discount factor | 0.99 |
| | Target network update rate | 0.005 |
| | Actor learning rate | 0.0004 |
| | Critic learning rate | 0.001 |
| Training | Start time steps | 15,000 |
| | Max time steps | 100,000 |
| | No. of repeated runs | 10 |
| | Evaluation frequency | 2,500 |
| | Episodes per evaluation Step | 6 |

All parameters not explicitly stated in the respective experiment sections are set to the default presented here.

**TABLE 2** Comparison of the total average runtimes for the $a_T$ and $h_T$ trials on InvertedPendulum.

| | Run | Mean total runtime |
|---|---|---|
| $a_T$ | Baseline | 54min |
| | $a_T = 0.98$ | 4h 2 min |
| | $a_T = 0.96$ | 3h 42 min |
| | $a_T = 0.92$ | 2h 12 min |
| | $a_T = 0.92$ | 1h 42 min |
| $h_T$ | $h_T = 0.9$ | 2h 54 min |
| | $h_T = 0.65$ | 2h 42 min |
| | $h_T = 0.45$ | 2h 12 min |
| | $h_T = 0.3$ | 2h 12 min |

## 5.2. Results

We ran two experiments for each environment comparing the effect of the activation and habituation thresholds, as shown in Figures 2, 3. Both thresholds allow for memory reduction and the modulation of node density by causing more states to be merged into nodes. For each experiment, we reduced one of the threshold values while keeping the respective other one at 1, to isolate their influence on node insertion. All experiments show a general decrease in performance as the memory size is reduced. However, the decline is much more drastic when using the habituation threshold.

TABLE 3 Comparison of the total average runtimes for the $a_T$ and $h_T$ trials on Reacher.

| | Run | Mean total runtime |
|---|---|---|
| $a_T$ | Baseline | 18min |
| | $a_T = 0.91$ | 4h 25min |
| | $a_T = 0.83$ | 4h 18min |
| | $a_T = 0.68$ | 2h 52min |
| | $a_T = 0.55$ | 1h 58min |
| $h_T$ | $h_T = 0.9$ | 4h 45min |
| | $h_T = 0.65$ | 3h 51min |
| | $h_T = 0.45$ | 3h 23min |
| | $h_T = 0.3$ | 3h 7min |

TABLE 4 Comparison of the total average runtimes for the $a_T$ and $h_T$ trials on HalfCheetah.

| | Run | Mean total runtime |
|---|---|---|
| $a_T$ | Baseline | 18min |
| | $a_T = 0.55$ | 6 h 31 min |
| | $a_T = 0.29$ | 5 h 42 min |
| | $a_T = 0.15$ | 4 h 38 min |
| | $a_T = 0.07$ | 3 h 44 min |
| $h_T$ | $h_T = 0.9$ | 3 h 6 min |
| | $h_T = 0.65$ | 2 h 32 min |
| | $h_T = 0.45$ | 2 h 14 min |
| | $h_T = 0.3$ | 2 h 1 min |

### 5.2.1. Activation threshold

For InvertedPendulum (Figure 2), lowering $a_T$ to 0.96 leads to a significant drop in memory size (Figure 2B). Although with an increased standard deviation, the performance stays close to that of the baseline, with only approximately 75% of baseline memory (Figure 2A). Setting $a_T = 0.92$ causes a decrease in performance that is, however, not proportional to the large decrease in memory size to only approximately 30% of the baseline's memory (Figure 2B). Here, the standard deviation becomes very large with some runs performing close to the baseline and some underperforming greatly. Going lower to approximately 15% of memory size with $a_T = 0.88$, we can observe a very sharp decrease in performance. When $a_T = 0.98$, the performance is almost identical to the baseline, which is expected as the memory size is roughly the same, leading to very similar sample batches.

The results are similar for the much more complex Half Cheetah when using much lower $a_T$-values. The values 0.55, 0.29, and 0.15 yield baseline performance (Figure 2E), with the latter two saving about 80% and 60% memories, respectively (Figure 2F); and the deviation staying similar. When memory size is reduced to approximately 40% with $a_T = 0.07$, the performance is worsened significantly with the standard deviation increasing continuously. This suggests a dynamic of some runs starting out well, due to favorable random initialization, and being able to progress steadily and gather useful data, whereas others fail to converge to successful policies and are unable to build up a comprehensive memory, a vicious cycle for the latter.

Reacher trials surprisingly do not seem to be affected by lowered memory at all. Neither the $a_T$-trials (Figure 2C) nor the $h_T$-trials diverge in performance (Figure 3C). Memory behaves somewhat logarithmically, as using the $a_T$-threshold causes the network to fill out the whole state space with nodes, leading to a converging behavior. Remarkably, convergence at approximately 20% of memory is achieved for Reacher with $a_T = 0.55$ (Figure 2D), with no impact on performance.

For the HalfCheetah runs, the lower the $a_T$, the higher the standard deviation for memory (Figure 2F). We assume a positive correlation between performance and memory size, as better performing runs reach farther in the HalfCheetah setup, experiencing more different parts of the state space (meaning more inserted nodes) than those that start out poor and just move back and forth at the initialized position.

For InvertedPendulum, the memory for $a_T = 0.98$ is for a short moment slightly higher than the baseline (Figure 2B), which is due to the different definitions of memory size, where the GWR-R increases its memory size by two at the start of each episode as it creates nodes for both the new start state and the next state. The baseline, on the other hand, only stores one tuple comprising both states. This effect is more pronounced at the start of the training when episodes are shorter as the agent is still poor at balancing the pendulum for long.

### 5.2.2. Habituation threshold

The results for the habituation threshold shown in Figure 3 stand in contrast to those for the activation threshold. While our tested values lead to different memory sizes, notably worse proportional performance becomes clear.

On InvertedPendulum, all runs approach the maximum reward of 1000 slowly, while only $h_T = 0.9$ and $h_T = 0.65$ approach closer to it at the end of training (Figure 3A). Although $h_T = 0.45$ and $h_T = 0.3$ have very comparable relative memory sizes to $a_T = 0.92$ of approximately 40% and 30%, respectively (Figure 3B), they have trouble surpassing even a reward of 400, with $h_T = 0.3$ declining in performance at the end of training. There is a very minor difference in memory size between $h_T = 0.65$ and $h_T = 0.45$ runs, but a stark difference in performance. For HalfCheetah, the results are even more unfortunate, with all values leading to significantly reduced performance and never growing beyond a maximum of 1500 reward (Figure 3E). Similar to the previous experiment, Reacher is again not impacted by reduced memory (Figure 3C).

### 5.2.3. Threshold comparison

We summarize the comparison of the two thresholds in Figure 4, which plots the averaged reward over all timesteps against the total memory at the end of the run on InvertedPendulum (Figure 4A), Reacher (Figure 4B), and HalfCheetah (Figure 4C). Here, the performance differences between the two thresholds become apparent with the curve for $h_T$ declining more steeply in reward collected in proportion to reduced memory size.

The use of the GWR-R unfortunately yields heavily increased learning times, as the distance calculation for finding the BMU takes longer, the network grows larger. This reaches from 2x to more than 22x the required training time, as shown in Tables 2–4.

## 6. Discussion

The results do not confirm our hypothesis that our approach causes more stable learning. The learning stability does not improve with reduced memory, it rather becomes poor. In general, the more compressed the memory, the poor the performance we obtain, but this does not happen in proportion to the decrease in memory. It seems that a good reduction in memory of approximately 20% to 40% is feasible without impacting the performance too much. Furthermore, it appears that a reduction in memory through the use of the activation threshold yields a minor decrease in performance than when using the habituation threshold.

We did not expect to see this result, as the habituation counter is a common feature of the GWR models, but when introducing temporal relations as we do and connecting nodes from non-independent and identically distributed (non-I.I.D) data that is highly correlated, merging samples into nodes in this way seems to be a problem. We theorize that one reason for decreases in performance with lower memory size is that the amount of incorrect tuples in the replay buffer (or rather their deviation from reality) rises due to averaging. When merging two consecutive samples into two existing nodes, the way nodes and edges are updated is different. Nodes are moved by some learning rate, while the action and reward are averaged with a moving average, with fresher samples weighting more. This will cause the stored action to deviate from the real action that connects the stored states in the environment, indicating that using an action stored along an edge between $s$ and $s'$ would in reality not lead from $s$ to $s'$, but only to a state near $s'$. As we are in the continuous domain, however, the action should always still be possible and because of the way the model is constructed, the maximum distance between the state one would actually end up in when using $a$ from $s$ and the stored $s'$ (the updated center of the node that best matches $s'$) is influenced by the thresholds, although a specific threshold choice seems to be vital. When decreasing the memory size through the activation threshold, the error along edges is somewhat limited as all actions integrated do originate and end at least in the vicinity of the node centers. The error will increase when lowering the activation threshold as this will lead to a greater region a node is responsible for. If the activation threshold is too low, the maximum distance from the node to a sample will increase, increasing the potential error along the node's edges. It would also cause more merging in general, which would lead not only to larger but generally more such errors over all the stored edges (as they stem from the merging process).

When using the habituation threshold, the potential of such errors is magnified. As the habituation threshold allows for samples to be merged irrespective of their distance to the BMU, as long as there is no closer node, the matching action and state values might further and further drift apart to something unrepresentative of the actual environment dynamics.

A related problem is that nodes are often part of multiple trajectories (e.g., the trajectory from node 1 to node 2 and then to node 3, as shown in Figure 1). If a new trajectory is learned and nodes change positions, existing edges from other trajectories are not updated. This will lead the stored actions along those edges to deviate more and more from the actual action that would connect the moved node to its neighbors in the trajectory. The habituation threshold allows very drastic changes in a position toward far away samples which will render parts of the established trajectories virtually unusable. Only by altering the activation threshold, edge errors, produced in this way, are again limited in proportion to the maximum distance from the node center.

In the InvertedPendulum habituation trial, it is interesting to note, however, the slow but steady performance increase of runs shown in green and orange (Figure 3A). We explain this as follows, While the habituation counter does allow samples to theoretically be arbitrarily far away from the BMU, the BMU is still the best-matching-unit, that is, the closest node to the sample of all the nodes. With increasing memory size and the map space being filled out more and more, the average distance between the sample and the BMU should decrease so the potential damage made by the habituation threshold will become smaller and new transitions will become more accurate – although the old, errorful ones can still be sampled and will not be corrected, continuing to play a disruptive role. The HalfCheetah environment on the other hand has state and action spaces that are too large for this to play a role in the limited training time we have set. It should follow that the harmful role played by the habituation threshold will also increase errors when combined with a lowered activation threshold, and we conclude that reducing memory size only through the activation threshold will lead to a minor decrease in performance proportional to memory.

The experiments we ran to explore the influence of $\phi$, the parameter controlling the decay of old samples in the moving average for network edges, are not shown as we found little to no influence on performance. A $\phi$ value of 1 might increase instability as it would increase the above-mentioned error where the action values move further toward a new sample than the state values, so it is probably best to keep $\phi$ somewhere in between the extremes. We would theoretically recommend something close to 0.1.

A thorough investigation of the actual extent of the above-suspected errors, comparing different thresholds and values, might give better intuitive insight into ways to mitigate them. A way to avoid learning from errorful samples might be to use the GWR-R's structure only to guide the selection of representative samples while not merging any values. Some further experiments and analyses are needed to investigate the source of the memory advantage and the ways to implement it in a more concise form.

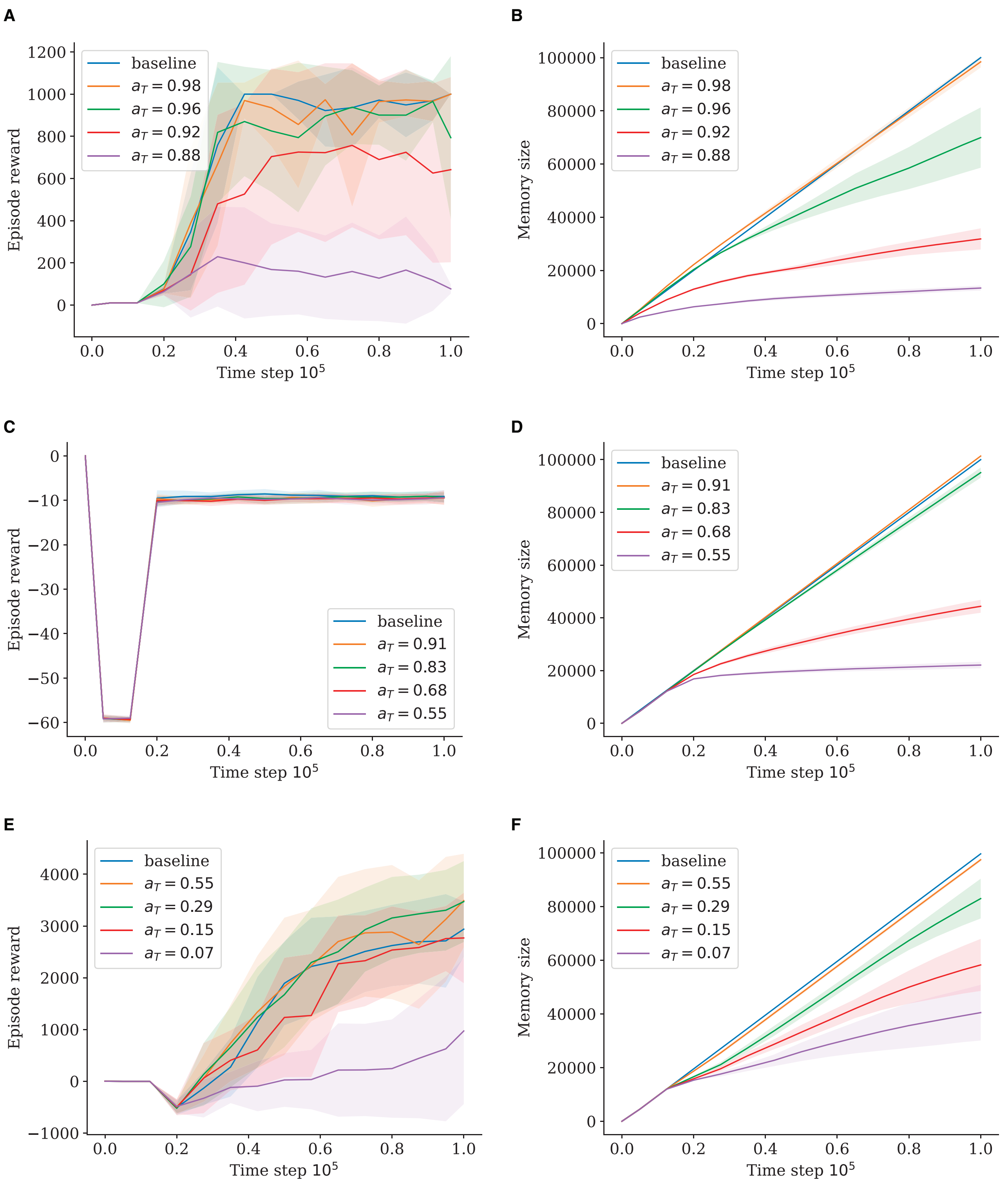

FIGURE 2
Episode reward and memory size for different activation threshold values ($h_T = 1$). The shaded area shows standard deviation. **(A)** Episode Reward for InvertedPendulum. **(B)** Memory Size for InvertedPendulum. **(C)** Episode Reward for Reacher. **(D)** Memory Size for Reacher. **(E)** Episode Reward for HalfCheetah. **(F)** Memory Size for HalfCheetah.

## 7. Conclusion

We propose a novel approach for replacing standard RL replay buffers, which store state transitions as they are, with a more semantic, world model-like approach using a modified GGWR (Parisi et al., 2017) architecture, the GWR-R, which merges similar states into nodes and stores state transitions along network edges.

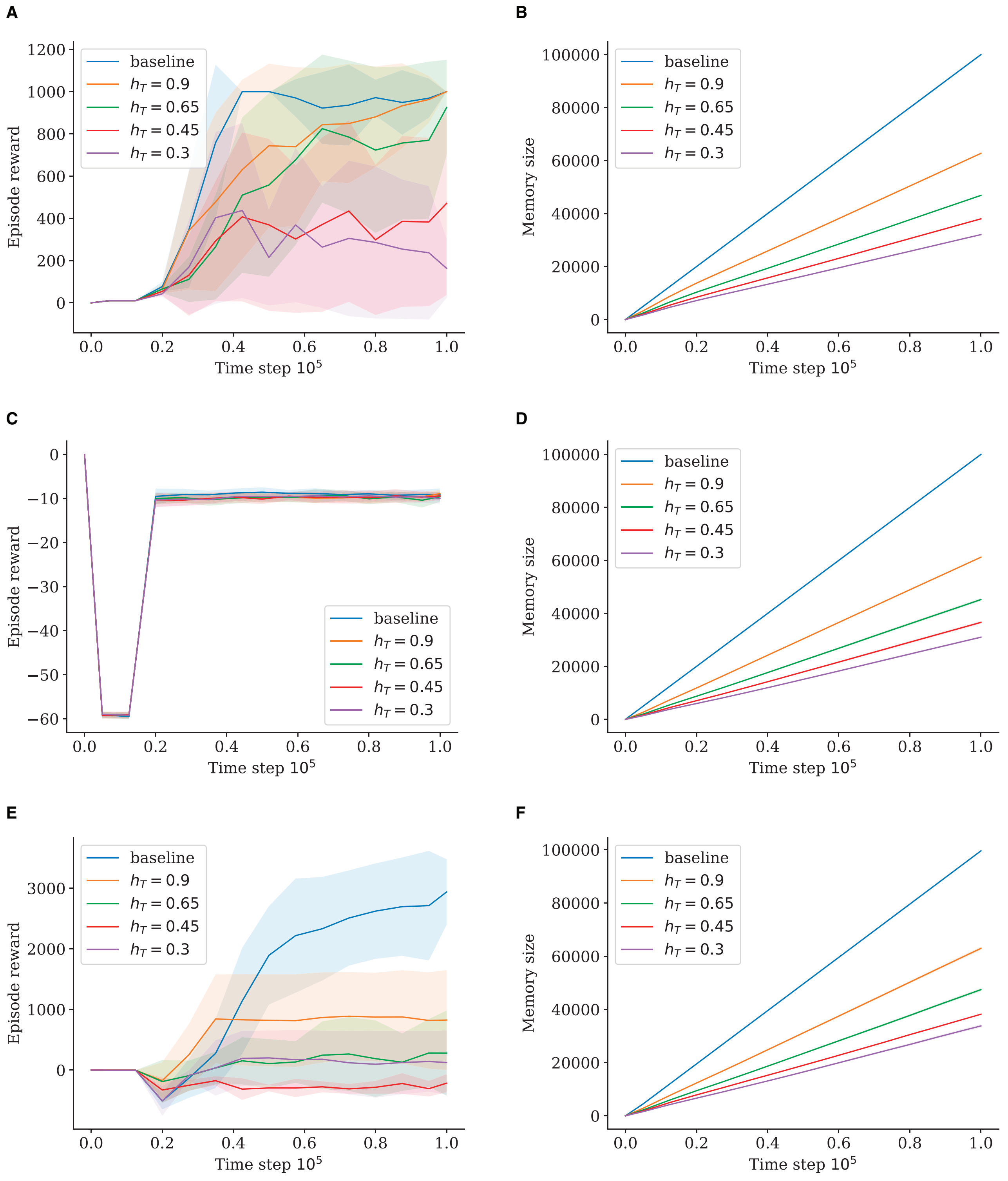

FIGURE 3
Episode reward and memory size for different habituation threshold values ($a_T = 1$). The shaded area shows standard deviation. **(A)** Episode Reward for InvertedPendulum. **(B)** Memory Size for InvertedPendulum. **(C)** Episode Reward for Reacher. **(D)** Memory Size for Reacher. **(E)** Episode Reward for HalfCheetah. **(F)** Memory Size for HalfCheetah.

Only decreasing the activation threshold that modulates the proportion of node insertion and node merging by changing the size of the merge area of a node, we were able to significantly reduce the amount of distinctly stored samples by 40—80% with minor to disproportionately small performance losses compared to the baseline on four different MuJoCo environments. We found that decreasing the memory size through the habituation threshold yields a sharper decline in performance. Our study opens up many questions and avenues to continue research for more compact, semantic replay memories (see Section 7.1). One limitation of our approach is that it requires additional computational time for constructing and adapting the growing self-organizing network necessary for experience replay. Overall, our study shows that map-based experience replay allows for significant memory reduction with only small decreases in performance. We suggest that our findings can make an important step toward efficiently solving catastrophic forgetting in RL, contributing to the development in the emerging research direction of continual robot learning.

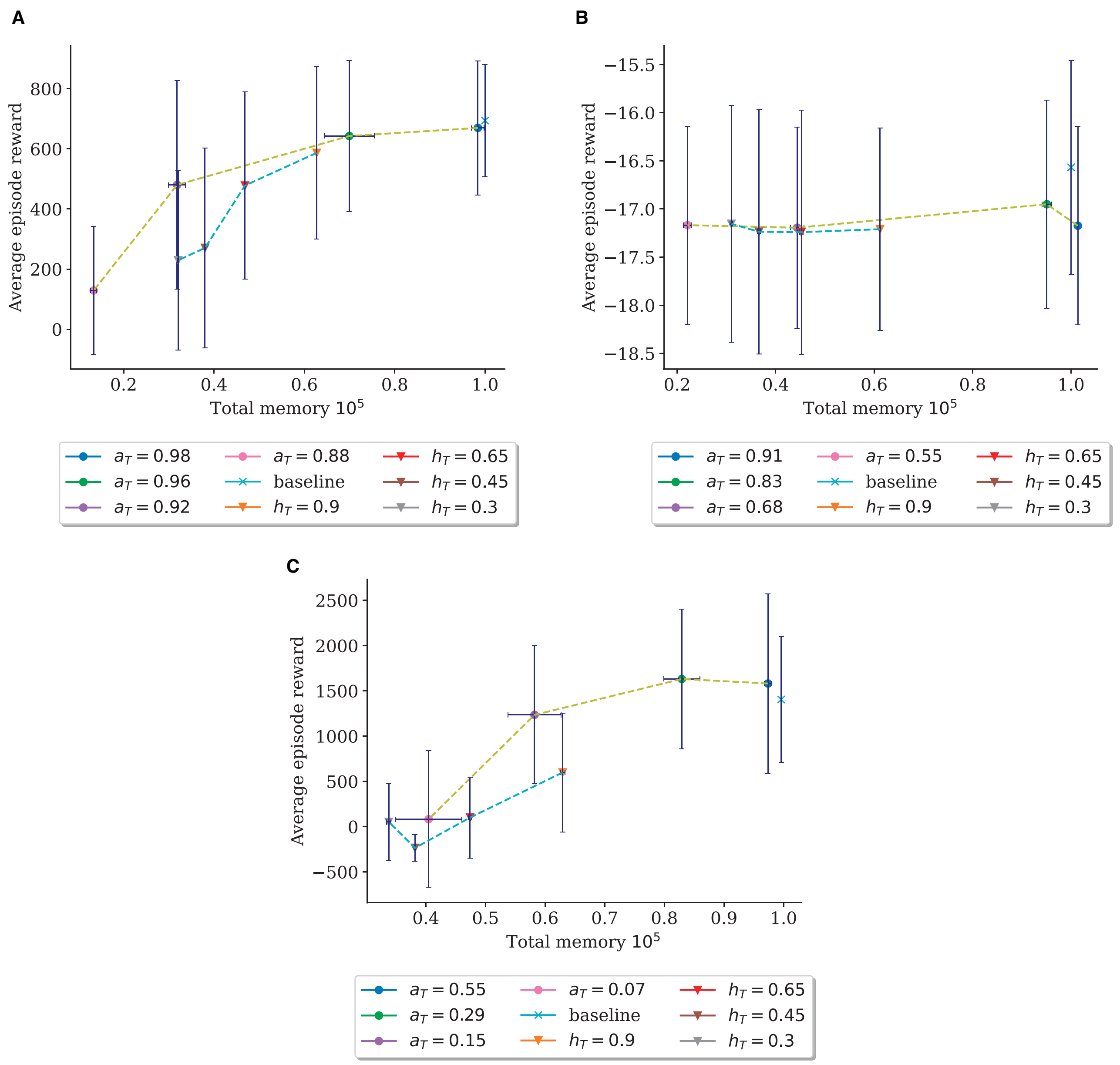

FIGURE 4
Comparison between the $a_T$-trial (dots) and the $h_T$-trial (triangles) of the ratio of average reward over all episodes to the memory size after training. Each run is one dot/triangle; dots/triangles are connected for easier comparison of approximate underlying functions. Error bars show the mean standard deviation. **(A)** Threshold Comparison for InvertedPendulum. **(B)** Threshold Comparison for Reacher. **(C)** Threshold Comparison for HalfCheetah.

## 7.1. Future work

Due to its demonstrated potential in overcoming catastrophic forgetting, we plan to extend our approach to address vision-based RL in more complex robotic scenarios, such as learning human-like gestures in our NICO robot (Kerzel et al., 2017). To enable this, our GWR-R approach can be used to generalize over raw pixels and learn a representation space with a graph structure that facilitates generating more diverse experiences, potentially speeding up learning control policies from pixels. Another direction for future work is to train our approach in continual multi-task RL settings, which we hypothesize could greatly benefit from our incrementally learned transition map. For example, our approach can replay experiences of old and new tasks easily by traversing the temporal edges built between visited states of different tasks, which would otherwise be difficult to perform with standard experience replay. Additionally, training our approach from multimodal sensory feedback (e.g., sound and vision Zhao et al., 2022) is an interesting direction for future work.

## Data availability statement

The raw data supporting the conclusions of this article will be made available by the authors, without undue reservation.

## Author contributions

All authors listed have made a substantial, direct, and intellectual contribution to the work and approved it for publication.

## Funding

This research was partially supported by the Federal Ministry for Economic Affairs and Climate Action (BMWK) under the project VeriKas and the German Research Foundation (DFG) under the projects Transregio Crossmodal Learning (TRR 169) and MoReSpace.

## Conflict of interest

The authors declare that the research was conducted in the absence of any commercial or financial relationships that could be construed as a potential conflict of interest.

## Publisher's note

All claims expressed in this article are solely those of the authors and do not necessarily represent those of their affiliated organizations, or those of the publisher, the editors and the reviewers. Any product that may be evaluated in this article, or claim that may be made by its manufacturer, is not guaranteed or endorsed by the publisher.

## Supplementary material

The Supplementary Material for this article can be found online at: https://www.frontiersin.org/articles/10.3389/fnbot.2023.1127642/full#supplementary-material